# Stable Segmentation of Digital Image

M. Kharinov
St. Petersburg Institute for Informatics and Automation of Russian Academy of Sciences (SPIIRAS)
khar@iias.spb.su

## Аннотация

In the paper the optimal image segmentation by means of piecewise constant approximations is considered. The optimality is defined by a minimum value of the total squared error or by equivalent value of standard deviation of the approximation from the image. The optimal approximations are defined independently on the method of their obtaining and might be generated in different algorithms. We investigate the computation of the optimal approximation on the grounds of stability with respect to a given set of modifications. To obtain the optimal approximation the Mumford-Shuh model is generalized and developed, which in the computational part is combined with the Otsu method in multi-thresholding version. The proposed solution is proved analytically and experimentally on the example of the standard image.

## 1. ВВЕДЕНИЕ

В литературе проблему сегментации часто относят к слабо формализованным, что справедливо, если сегментация определяется не как результат, а как процесс разбиения пикселей изображения на множества по некоторому алгоритму. В этом случае, результирующее разбиение зависит от особенностей входных данных, что приводит к разделению изображений на ряд предметных областей (телевизионных, аэрокосмических, медицинских и пр.), для которых разрабатываются конкретные алгоритмы сегментации, а достоверность сегментации оценивается эмпирически по адекватности зрительному восприятию или результатам автоматического обнаружения заранее заданных объектов.

В более точной постановке проблема сегментации состоит в определении сегментированного изображения независимо от конкретного алгоритма или технологии его получения. В этом случае, вычисляемые сегменты отличаются от произвольных сегментов тем, что составляют оптимальные приближения изображения, минимально отличающиеся от изображения по значению некоторого функционала. При этом достоверность сегментации можно проверить формально без использования понятия зрительно воспринимаемых объектов, а выделение объектов трактовать как неформальную интерпретацию сегментации. В терминологии кластерного анализа [15], речь идет математической постановке задачи автоматической классификации, при которой искомые разбиения множеств на классы характеризуются оптимальным качеством. Одной из классических оценок качества разбиения является суммарная квадратичная ошибка или взвешенная сумма внутриклассовых дисперсий, которая является основой метода $k$–средних [6,7,15] и ряда других методов в абстрактном кластерном анализе и наглядной автоматической сегментации цифровых изображений.

## 2. ПОСТАНОВКА ЗАДАЧИ

Классическая оценка качества разбиения интенсивно применяется в методах оптимальной сегментации, однако в литературе приводятся недостаточно полные данные по оптимальным приближениям даже для стандартных изображений, в частности, для изображения «Лена». В гистограммном методе Оцу [11] строится оптимальное разбиение пикселей изображения на два множества (класса), отвечающее минимальному значению обсуждаемого функционала в одномерном варианте. В известной работе [5], для задач выделения на изображении обычных и текстурных объектов, реализуется мультипороговый метод Оцу и предпринимается попытка вычисления последовательности оптимальных приближений изображения. Результаты, однако, ограничиваются начальными пятью приближениями изображения из-за экспоненциального роста продолжительности вычислений с увеличением числа классов (градаций яркости), на которые разбивается множество пикселей изображения.

В случае последовательного разбиения пикселей по яркости на $m=1,2,...$ градаций, минимизируемый функционал $\sigma^2$ записывается как взвешенная сумма дисперсий $\sigma_i^2$ по множествам $i$:

$$\sigma^2 = \sum_{i=1}^{m} \omega_i \sigma_i^2, \qquad (1)$$

где $\omega_i$ — вероятность, сопоставляемая множеству $i$ (в [5] рассматривается эквивалентный функционал).

Для изображения из $N$ пикселей величина $N\sigma^2$ есть сумма квадратов отклонений яркостей пикселей от средних значений, вычисленных по каждому множеству пикселей. В модели Мамфорда–Шаха [1,4,8,9,12], в которой рассматриваются множества пикселей связных сегментов изображения, эта величина является основной компонентой «энергетического» функционала [4]:

$$N\sigma^2 + \lambda L, \qquad (2)$$

где $L$ — суммарная длина границ между сегментами, а $\lambda$ — так называемый, «регуляризационный» параметр.

Величина $\sigma$ в (2) описывает среднеквадратичное отклонение изображения от своего кусочно–постоянного приближения с усредненными по сегментам значениями пикселей.

В популярной работе [4] ряд последовательных значений параметра $\lambda$ устанавливается эвристически в процессе обработки изображения. В современной практической версии модели FLSA (Full $\lambda$–Scheduled Algorithm, [1,12]), применяемой в программном комплексе "ENVI" (Environment for Visualizing Images, http://www.ittvis.com/portals/0/pdfs/envi/Feature_Extraction_Module.pdf), параметр $\lambda$ устанавливается автоматически из условия неизменности функционала (2) при слиянии сегментов, что нарушает идею минимизации функционала, как справедливо отмечается в [1]. В [16] указанный параметр полагается равным 0, что не противоречит (2), и сохраняет нетривиальный смысл минимизации $\sigma$, если она выполняется для каждого числа мизации $\sigma$, если она выполняется для каждого числа сегментов. В этом случае, оптимальность приближений определяется в классическом смысле — минимальным отличием приближения от изображения по среднеквадратичному отклонению $\sigma$ или суммарной квадратичной ошибке $E \equiv N\sigma^2$:

$$\sigma = \min \quad \Leftrightarrow \quad E = \min, \quad (3)$$

что, в принципе, не исключает возможности учета границ между сегментами посредством аддитивной или мультипликативной добавок в критерии слияния, как в версии [4] или версии FLSA [1,12] модели Мамфорда–Шаха.

Как мы установили в предыдущей работе [2], принципиальное ограничение вычислительной схемы в модели Мамфорда–Шаха заключается в недостаточности операции слияния сегментов для получения, вообще говоря, неиерархической последовательности оптимальных приближений. Для преодоления указанного недостатка мы обобщаем операцию слияния и вводим дополнительную операцию коррекции множеств пикселей изображения.

## 3. АНАЛИТИЧЕСКОЕ ОБОБЩЕНИЕ ОПЕРАЦИИ СЛИЯНИЯ

Рассмотрим реклассификацию $k$ пикселей средней яркости $I$, исключаемых из множества 1 пикселей со средней яркостью $I_1$ и включаемых в число пикселей множества 2 со средней яркостью $I_2$. Сопутствующее приращение $\Delta E$ квадратичной ошибки $E$ выражается в виде:

$$\Delta E = \begin{cases} \Delta E_{merge}, & k = n_1, \\ \Delta E_{correct}, & k < n_1, \end{cases} \quad (4)$$

где $n_1, n_2$ — число пикселей в множествах 1 и 2, соответственно, а $\Delta E_{merge}$ и $\Delta E_{correct}$ имеют вид:

$$\Delta E_{merge} = \frac{(I_1 - I_2)^2}{\frac{1}{n_1} + \frac{1}{n_2}}, \quad \Delta E_{correct} = \frac{(I - I_2)^2}{\frac{1}{k} + \frac{1}{n_2}} - \frac{(I - I_1)^2}{\frac{1}{k} - \frac{1}{n_1}}. \quad (5)$$

$\Delta E_{merge}$ описывает приращение квадратичной ошибки при слиянии множеств пикселей посредством реклассификации всех пикселей из множества 1 в множество 2 и уменьшении числа множеств на 1. $\Delta E_{correct}$ описывает приращение квадратичной ошибки при реклассификации части пикселей из множества 1 в множество 2 и сохранении числа множеств. Далее условимся разделять термины «слияние» и «реклассификация», употребляя последний только для обозначения модификации классификации части пикселей.

В вычислениях по Мамфорду–Шаху и [16] сегментация выполняется посредством итеративного слияния рассматриваемых множеств пикселей. Следуя [16], слияние множеств 1 и 2 выполняется при минимальной величине приращения квадратичной ошибки $\Delta E_{merge}$:

$$\Delta E_{merge} = \min. \quad (6)$$

Коррекция множеств выполняется посредством реклассификации пикселей из условия уменьшения квадратичной ошибки. Критерием коррекции является отрицательная величина приращения квадратичной ошибки $\Delta E_{correct}$:

$$\Delta E_{correct} < 0. \quad (7)$$

Элементарным преобразованием из (5) и (7) получаем условие коррекции, учитывающее только знак $\Delta E_{correct}$:

$$|I - I_1| > \alpha \cdot |I - I_2|, \quad (8)$$

где коэффициент $\alpha < 1$ описывает соотношение количеств пикселей в множествах и определяется в виде:

$$\alpha \equiv \sqrt{\frac{n_2(n_1 - k)}{n_1(n_2 + k)}}. \quad (9)$$

Из (8), (9) для реклассификации $k$ одинаковых пикселей следует, что если среднеквадратичное отклонение снижается при реклассификации одного пикселя, то оно снижается и при реклассификации остальных пикселей той же яркости, поскольку с ростом $k$ коэффициент $\alpha$ уменьшается.

Применяя логическое отрицание к выражениям (7), (8), получаем условие устойчивости:

$$\Delta E_{correct} \geq 0 \quad \Leftrightarrow \quad |I - I_1| \leq \alpha \cdot |I - I_2|. \quad (10)$$

Разбиение пикселей изображения на множества считается устойчивым, если для всех рассматриваемых пар множеств реклассификация заданных подмножеств пикселей из одного множества в другое не приводит к уменьшению $\sigma$.

Очевидно, что оптимальное разбиение изображения является устойчивым, но устойчивое разбиение не обязательно оптимально. Ограничения на допустимые пары множеств 1 и 2, например, обменивающихся пикселями при сохранении связности, или рассмотрение одновременной реклассификации двух и более подмножеств пикселей приводит к различным вариантам программной реализации условия устойчивости (10). Здесь же следует заметить, что метод $k$-средних [6,7,15], вообще говоря, приводит к неустойчивому разбиению пикселей с завышенным значением $\sigma$, поскольку в этом методе не учитывается коэффициент $\alpha$ (9).

## 4. МЕТОДЫ МИНИМИЗАЦИИ $\sigma$

Для вычисления последовательности оптимальных разбиений пикселей изображения на множества, обеспечивающих минимизацию среднеквадратичного отклонения приближений от изображения, мы воспроизвели вычисления в версии FLSA модели Мамфорда-Шаха и, для сравнения, в версии [16]. Затем, в рамках версии [16], реализовали вычисление неиерархических приближений с обычными связными сегментами и со связными множествами из несвязных пикселей, а также разработали новую версию мультипорогового неиерархического метода Оцу.

В мультипороговом методе Оцу искомые множества пикселей отвечают диапазонам яркости, на которые по гистограмме разбивается рабочий диапазон яркости. При этом учитывается свойство компактности оптимальных разбиений по яркостной оси, состоящее в том, что число классов, на которые разбивается множество пикселей, совпадает с числом диапазонов (градаций), на которые разделяется рабочий диапазон яркости. Требуется разбить рабочий диапазон яркости на последовательное число диапазонов так, чтобы при замещении значений пикселей усредненными по диапазонам значениями яркости, среднеквадратичное отклонение $\sigma$ полученного приближения от изображения оказалось минимальным при каждом числе диапазонов от 1 до максимального значения $M$, равного числу встречающихся на изображении яркостей. В отличие от [5], в нашей реализации метода Оцу строится полная последовательность оптимальных разбиений.

Пара тривиальных разбиений на 1 и на $M$ диапазонов задана. Остальные разбиения получаются и преобразуются слиянием или дроблением текущих диапазонов с изменением числа диапазонов и числа соответствующих градаций средней яркости на единицу, как в [6]. Для каждого разбиения итеративно осуществляется минимизация квадратичной ошибки $E$ «по частям», которая выполняется для всех перекрывающихся участков гистограммы, состоящих не более чем из $l$ последовательных диапазонов яркости ($l = 2,3,...$). При этом оптимальные пороговые значения яркости находятся перебором вариантов.

Эксперимент показал, что с ростом $l$ возрастает продолжительность обработки, как в [5]. Но, с другой стороны, уже при начальных значениях $l$ последовательность искомых приближений стабилизируется, и снижается максимальная величина модификации $\sigma$, нулевое или близкое к нулю значение которой может служить критерием окончания оптимизации. Расчет устойчивых приближений и их оптимизация по $\sigma$ выполняется по формулам (4)-(10) без суммирования квадратов значений яркости.

В разработанном обобщенном варианте вычислений по модели Мамфорда–Шаха и [16] в качестве начального берется любое разбиение пикселей изображения на $N$ множеств, в частности, на отдельные пиксели, которое затем модифицируется за счет операций слияния и коррекции. Строится последовательность разбиений от 1 до $N$. Для множеств пикселей задается отношение смежности по аналогии со связными сегментами — два множества считаются смежными, если на изображении встречаются смежные пиксели из этих множеств. Множества пикселей делятся на подмножества, образуемые пикселями одинаковой яркости. В качестве указанных подмножеств рассматривались либо отдельные пиксели, либо все пиксели каждой яркости из числа пикселей, отнесенных к данному множеству.

Минимизация среднеквадратичного отклонения $\sigma$ приближений от изображения выполняется в алгоритмах поочередной коррекции разбиений пикселей изображения на множества и слияния смежных множеств пикселей.

Слияние смежных множеств пикселей выполняется из условия (6) и сопровождается уменьшением числа множеств пикселей, а также увеличением $\sigma$. В процессе коррекции смежные множества пикселей обмениваются подмножествами пикселей одинаковой яркости из условий (7), (8) уменьшения $\sigma$. Из нескольких возможностей реклассификации данного подмножества пикселей выбирается та, что обеспечивает максимальное падение $\sigma$ согласно выражению $\Delta E_{correct}$ в (5). Коррекция циклически повторяется до получения устойчивого разбиения изображения на множества пикселей, при котором реклассификация любого предусмотренного подмножества пикселей из данного множества в смежное увеличивает среднеквадратичное отклонение, или, по крайней мере, оставляет его неизменным. На каждом шаге коррекции среднеквадратичное отклонение падает, а число множеств пикселей не меняется.

Результат сегментации зависит от условий выполнения коррекции множеств пикселей. При блокировании коррекции, нарушающей связность пикселей, вычисляется последовательность разбиений изображения на связные сегменты. Если же при коррекции допускается нарушение связности сегмента–донора, то результатом сегментации служат множества пикселей в виде наборов несмежных сегментов с одинаковой средней яркостью.

## 5. ЭКСПЕРИМЕНТАЛЬНЫЕ РЕЗУЛЬТАТЫ

Рис. 1 иллюстрирует оптимальные приближения, вычисленные для стандартного изображения «Лена».

$\sigma = 31.60341$

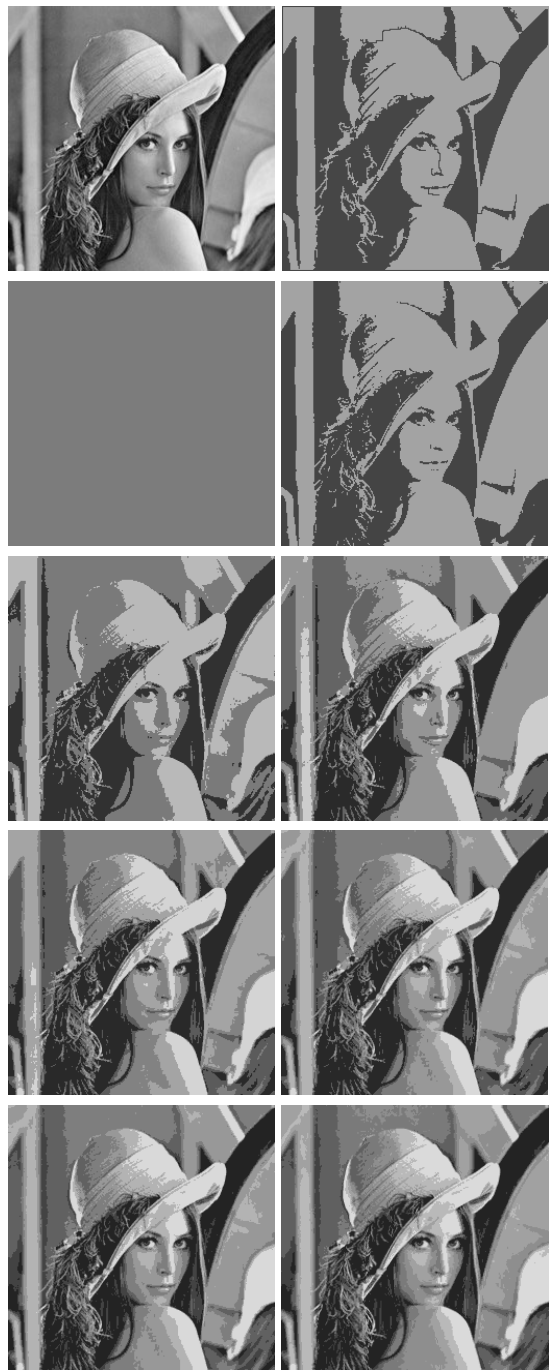

**Рисунок 1:** Оптимальные приближения изображения.

Исходное изображение показано в левом верхнем углу. Рядом, в правом верхнем углу, показано «близкое к оптимальному» приближение исходного изображения двумя связными сегментами [2], — значение среднеквадратичного отклонения $\sigma$ указано над приближением. Остальные 8 изображений иллюстрируют оптимальные приближения исходного изображения в последовательном числе градаций яркости и размещены в порядке возрастания числа градаций от 1 до 8 слева направо и сверху вниз.

Нетрудно заметить, что приближение изображения двумя связными сегментами имеет выраженную структуру и состоит из «площадок» и «соединительных» элементов шириной в один пиксель, слабо влияющих на среднеквадратичное отклонение. При этом оптимальное приближение в двух градациях можно преобразовать в приближение, близкое к оптимальному, если отдельные «площадки» на нем стереть, а остальные соединить между собой посредством «соединительных» элементов.

Всего оптимальных приближений изображения в различном числе градаций 216, так как в изображении «Лена» встречаются пиксели 216 градаций исходной яркости. Для приближений в $m = 1, 2, \ldots, 48$ усредненных яркостных градациях значения $\sigma$ перечислены в четных столбцах таблицы — по 16 штук в каждом столбце.

Оптимальные приближения вычислены «по частям» в нашей реализации мультипорогового метода Оцу при $l \leq 3$. Проверено, что на полученные разбиения не влияет преобразование изображения в негатив, хотя процесс вычисления приближений при этом меняется. В ходе экспериментов ни одно перечисленных в таблице значений $\sigma$ не удалось уменьшить за счет применения алгоритмов анализа связных множеств и др. Однако пока мы не исключаем, что при следующих значениях $l$ некоторые значения $\sigma$ окажется возможным уточнить[1].

Характерно, что последовательность оптимальных приближений изображения не является иерархической. При этом яркостные диапазоны из различных разбиений гистограммы и сегменты из различных разбиений изображения перекрываются между собой (рис. 1).

Если в мультипороговом методе Оцу перебор пар множеств пикселей ограничивается благодаря условию компактности, то в вычислениях по модели Мамфорда-Шаха и [16] сравниваются только смежные множества пикселей. Интересно выяснить, как согласуются при этом результаты расчетов в той и другой моделях.

Для изучения эффекта учета пространственного распределения пикселей мы выполнили программную реализацию модели Мамфорда–Шаха, поддерживающую операции слияния и коррекции для любых множеств пикселей. При начальном разбиении изображения на связные сегменты и единственной операции слияния смежных сегментов выполняются вычисления со связными сегментами, как в классических вариантах модели Мамфорда–Шаха [1,4,8,9,12] и версии [16]. Операция коррекции, в общем случае, выводит из множества связных сегментов. Поэтому при выполнении коррекции связность пикселей, отнесенных к одному множеству, учитывается как самостоятельное условие [2], или игнорируется. В последнем случае, учет геометрических свойств множеств пикселей ограничивается отношением смежности между множествами.

Зависимость $\sigma$ от числа множеств пикселей для результирующих разбиений иллюстрируется графиками на рис. 2.

На рис. 2 зависимость оценок $\sigma$ от числа градаций яркости в диапазоне $m$ от 1 до 216 показана сплошной нисходящей линией, которая описывает результаты вычислений в нашей версии мультипорогового метода Оцу. Сливающаяся с

---

[1] Примеры оптимальных, точнее говоря, предельно близких к оптимальным приближений реальных изображений будут выложены на сайте СПИИРАН по электронному адресу http://www.spiiras.nw.ru с целью применения в задачах тестирования и сравнения алгоритмов оптимизации по $\sigma$.

**Таблица.**

**Оценки $\sigma$ для оптимальных приближений изображения «Лена» в $m$ градациях яркости**

| $m$ | $\sigma$ | $m$ | $\sigma$ | $m$ | $\sigma$ |
|---|---|---|---|---|---|
| 1 | 55.88322 | 17 | 3.70326 | 33 | 1.72854 |
| 2 | 30.64564 | 18 | 3.50441 | 34 | 1.68547 |
| 3 | 21.21739 | 19 | 3.32383 | 35 | 1.64729 |
| 4 | 14.96450 | 20 | 3.15658 | 36 | 1.61313 |
| 5 | 11.69762 | 21 | 3.01260 | 37 | 1.57894 |
| 6 | 10.03975 | 22 | 2.87305 | 38 | 1.54589 |
| 7 | 8.46072 | 23 | 2.73844 | 39 | 1.51218 |
| 8 | 7.51121 | 24 | 2.60572 | 40 | 1.47850 |
| 9 | 6.81359 | 25 | 2.49076 | 41 | 1.44586 |
| 10 | 6.14397 | 26 | 2.37391 | 42 | 1.41400 |
| 11 | 5.57864 | 27 | 2.25584 | 43 | 1.38327 |
| 12 | 5.11403 | 28 | 2.15389 | 44 | 1.35340 |
| 13 | 4.75689 | 29 | 2.05655 | 45 | 1.32287 |
| 14 | 4.42306 | 30 | 1.95565 | 46 | 1.29542 |
| 15 | 4.17825 | 31 | 1.87573 | 47 | 1.26794 |
| 16 | 3.92460 | 32 | 1.79485 | 48 | 1.23991 |

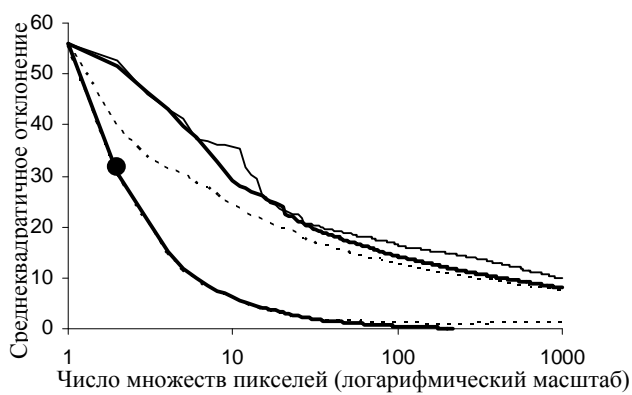

**Рисунок 2:** Графики зависимости оценок $\sigma$ от числа множеств пикселей для приближений изображения «Лена».

ней пунктирная кривая описывает результаты получения оптимальных разбиений изображения на 1, 2, ..., 1000 множеств пикселей в модели Мамфорда–Шаха без учета связности пикселей и начальном разбиении изображения на отдельные пиксели. Три верхних кривых описывают разбиения на связные сегменты. Пара переплетающихся кривых описывает иерархические приближения, полученных слиянием сегментов — с учетом границ между сегментами по версии FLSA [1,12] (верхняя тонкая кривая) и без учета границ (нижняя жирная кривая). Следующая пунктирная кривая описывает результаты автоматической генерации последовательности приближений изображения связными сегментами, полученные посредством чередования слияния сегментов и коррекции разбиений изображения на связные сегменты при условии сохранения числа сегментов в процессе коррекции [2]. Жирная точка, которая несколько смещена вверх относительно нисходящей линии оценок оптимальных значений $\sigma$, описывает близкое к оптимальному приближение изображения двумя связными сегментами (рис. 1), полученное автоматизированным преобразо-ванием оптимального приближения в двух градациях яркости с использованием интерактивного управления процессом установления «соединительных элементов».

Как видно на рис. 2, введение операции коррекции заметно улучшает приближения изображения по $\sigma$, выражающееся в том, что пунктирная кривая располагается ниже переплетающихся кривых. Тем не менее, поскольку жирная точка на рис. 2 оказывается почти в полтора раза ниже соответствующего значения $\sigma$ на пунктирной кривой, можно заключить, что автоматическая аппроксимация изображения связными сегментами [2] все же приводит к завышенным значениям $\sigma$ из-за не решенной пока проблемы эффективного автоматического установления соединительных элементов между сегментами. Как показано в [2], из-за недостаточно точной аппроксимации по $\sigma$, в сегментированном изображении пропускаются те или иные визуально воспринимаемые объекты, причем результаты разбиения изображения на несколько сегментов зависят от начального разбиения изображения, особенностей вычислений и пр. Указанную трудность удается обойти за счет рассмотрения смежных множеств пикселей без учета связности пикселей из одного множества, т.е. за счет ослабления связности.

Эксперимент показывает что, если в качестве начального разбиения на множества берется разбиение пикселей изображения по градациям исходной яркости с учетом смежности множеств, но без учета связности пикселей из одного множества, то получаемые разбиения воспроизводят результаты вычислений, полученные перебором пороговых значений по методу Оцу. График зависимости $\sigma$ от числа множеств, вычисленных в алгоритме слияния/коррекции, сливается с графиком $\sigma$ в зависимости от числа градаций.

Практически к тем же самым разбиениям на перекрывающиеся множества приводят вычисления при начальном разбиении изображения на отдельные пиксели и достаточной редукции числа множеств пикселей. В случае стандартного изображения (рис. 1), слияние смежных множеств пикселей по очереди с коррекцией получаемых разбиений приводит к тому, что при снижении числа формируемых множеств до 30 все пиксели каждой градации исходной яркости оказываются отнесенными к одному и тому же множеству, что обеспечивает воспроизведение оптимальных приближений в 1, 2, ..., 30 градациях яркости. По визуальному восприятию, сравниваемые приближения совпадают между собой, так как при более чем 30 множествах пикселей (градациях средней яркости) заметным образом не отличаются от исходного изображения.

## 6. ПРИМЕНЕНИЕ

Оптимальные приближения интересны не только тем, что их можно получать по различным алгоритмам. Оказывается, что они устойчивы относительно изменения масштаба изображения и получаются сходными для изображений в различной форме представления (рис. 3).

На рис.3 показан результат покомпонентной сегментации цветового изображения "Лена" из 512×512 пикселей (слева) с последующим преобразованием в полутоновое представление (справа), посредством усреднения оптимальных приближений в трех градациях каждой из RGB-компонент.

Несмотря на то, что обсуждаемое изображение рис. 3 цветовое и по размерам в два раза превосходит изображение рис. 1, результат его сегментации (справа на рис. 3) оказывается близок к результату сегментации в четырех градациях изображения "Лена" из 256×256 пикселей (в центральном ряду справа на рис. 1). Таким образом, формальное условие устойчивости (10) обеспечивает устойчивость сегментации изображения в прикладном смысле.

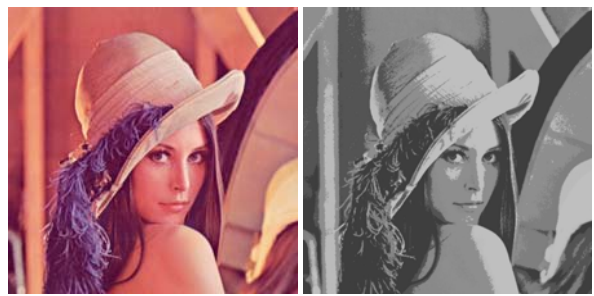

**Рисунок 3:** Сегментация цветового изображения.

В контексте проблемы автоматического распознавания смысл получения последовательности оптимальных приближений рис. 1 заключается в вычислении для выделения объектов множества сегментов, составляющих ничтожную долю всевозможных сегментов изображения. В этом случае, в процессе фильтрации при детектировании объектов необходимо проанализировать признаки сравнительно небольшого числа сегментов.

Особенностью оптимальных приближений изображения является то, что они задаются последовательностью разбиений с перекрытиями сегментов. При этом иерархическая структура оказывается признаком, который может использоваться для детектирования объектов.

## 7. ОСОБЕННОСТИ ВЫЧИСЛЕНИЙ

Обращаясь к модели Мамфорда-Шаха [1,12] и [16], не следует недооценивать вычислительной проблемы, связанной с необходимостью скоростного выполнения порядка $N-1$ итераций, где $N$ — число пикселей в изображении. Проблема усугубляется при встраивании в многочисленные итерации циклической же коррекции множеств пикселей, но решается без рутинной стандартной оптимизации программ за счет тысячекратного снижения числа итераций в быстром алгоритме одновременного парного слияния смежных множеств пикселей по всему полю изображения [2], который относится к алгоритмам поглощения [15].

С другой стороны, возникает проблема запоминания последовательности перекрывающихся разбиений на сегменты для оперативного преобразования и обработки в памяти компьютера. Эту проблему, вероятно, можно решить посредством аппроксимации последовательности перекрывающихся разбиений несколькими иерархическими последовательностями разбиений, для хранения, преобразования и фильтрации которых в фиксированном объеме памяти разрапботан аппарат дихотомической сегментации изображения в терминах структуры данных «динамических» деревьев Слейтора-Тарьяна [3,13,14,17]. Следует отметить, что упомянутый аппарат разрабатывается нами с начала 90-х годов прошлого века, тогда как в практику обработки изображений активно начинает внедряться только с начала текущего века [10], пока в менее освоенном варианте (http://www.lix.polytechnique.fr/~nielsen/Srmjava.java).

Во избежание зацикливания, при программировании коррекции разбиения изображения рекомендуется точно детектировать нулевое изменение $\Delta E_{correct}$ и обновлять значения параметров множеств пикселей (число пикселей и среднюю яркость) при каждой реклассификации их подмножеств. Следует обратить внимание также на то, что в режиме on-line модификации параметров множеств line модификации параметров множеств пикселей порядок анализа и преобразования множеств сказывается на процессе минимизации среднеквадратичного отклонения, но не влияет на результаты оптимизации, в случае достижения глобального минимума. При этом инвариантность результата сегментации может служить критерием окончания оптимизации, необходимого, например, при переборе пороговых значений яркости в мультипороговом методе Оцу. В противном случае, изменение порядка модификации множеств допустимо использовать как дополнительный прием для получения разбиений с минимальным среднеквадратичным отклонением $\sigma$.

Для исключения влияния изменения порядка слияния/коррекции множеств на процесс вычислений достаточно упорядочить пары множеств по величине изменения $\sigma$ или другому параметру. Однако, судя по первым экспериментам с различными изображениями, упорядочение пар множеств, существенно не влияет на результаты оптимизации, замедляя при этом вычисления.

Решение вычислительных проблем систематизации, упорядочения и фильтрации результатов сегментации для удобства выделения конкретных объектов является интересным направлением дальнейших исследований.

## 8. ЗАКЛЮЧЕНИЕ

Таким образом, в статье аналитически и экспериментально обоснован подход к сегментации цифровых изображений посредством оптимальных кусочно–постоянных приближений, характеризующихся минимальными значениями среднеквадратичного отклонения приближения от изображения. В качестве основы вычислений использованы метод Оцу и модель Мамфорда–Шаха, которые объединены в единой постановке задачи согласно версии [16].

В нашей реализации метода Оцу разработан гистограммный способ вычисления последовательных оптимальных приближений изображения посредством коррекции "по частям" и изменений числа градаций на единицу. Для оптимизации приближений по среднеквадратичному отклонению предложено дополнить слияние множеств пикселей в модели Мамфорда–Шаха и [16] операцией коррекции (7)-(9) и применять обобщение понятия связности множеств пикселей, к которым либо предъявляется, либо не предъявляется условие связности пикселей, относимых к одному множеству. В качестве формального признака оптимального разбиения установлено свойство (10) устойчивости сегментации относительно предусмотренных вариантов реклассификации пикселей.

Одним из перспективных направлений дальнейших исследований является обобщение формул (4)-(10) на многомерный случай в контексте разработки и реализации метода кластеризации, альтернативного методу *k*-средних, который, вероятно, будет полезен для улучшения сегментации цветовых изображений.

## 9. ССЫЛКИ

### Об авторе

М.В. Харинов — старший научный сотрудник СПИИРАН. Его адрес: khar@iias.spb.su.